%% file: main.tex
\def\year{2022}\relax
\documentclass[letterpaper]{article} % DO NOT CHANGE THIS
\usepackage{aaai22}  % DO NOT CHANGE THIS
\usepackage{times}  % DO NOT CHANGE THIS
\usepackage{helvet}  % DO NOT CHANGE THIS
\usepackage{courier}  % DO NOT CHANGE THIS
\usepackage[hyphens]{url}  % DO NOT CHANGE THIS
\usepackage{graphicx} % DO NOT CHANGE THIS
\usepackage{natbib}  % DO NOT CHANGE THIS AND DO NOT ADD ANY OPTIONS TO IT
\usepackage{enumitem}
\usepackage{amsmath,amsthm}
\usepackage{caption} % DO NOT CHANGE THIS AND DO NOT ADD ANY OPTIONS TO IT
\DeclareCaptionStyle{ruled}{labelfont=normalfont,labelsep=colon,strut=off} % DO NOT CHANGE THIS
\usepackage{algorithm}
\usepackage{algorithmic}

\usepackage{newfloat}
\usepackage{listings}
\floatstyle{ruled}
\newfloat{listing}{tb}{lst}{}
\floatname{listing}{Listing}
\title{A Safety Assurable Human-Inspired Perception Architecture}
\author{
    Rick Salay and
    Krzysztof Czarnecki
    }
\affiliations{
    University of Waterloo\\
    rsalay@gsd.uwaterloo.ca, kczarnec@gsd.uwaterloo.ca
}

\input{macros}

\begin{document}

\maketitle

\begin{abstract}
Although artificial intelligence-based perception (\aip) using deep neural networks (DNN) has achieved near human level performance, its well-known limitations are obstacles to the safety assurance needed in autonomous applications. These include vulnerability to adversarial inputs, inability to handle novel inputs and non-interpretability. While research in addressing these limitations is active, in this paper, we argue that a fundamentally different approach is needed to address them. Inspired by dual process models of human cognition, where Type 1 thinking is fast and non-conscious while Type 2 thinking is slow and based on conscious reasoning, we propose a dual process architecture for safe \aip. 
% Perception tasks using AI (e.g., detecting cyclists during automated driving) is dominated by type 1 style approaches using ML and have achieved high levels of success. Despite this, there remain crucial obstacles to safety assurable perception, even for achieving a human baseline level of performance. In this paper, we argue for the position that a dual process approach to perception is needed in AI to address these obstacles. 
We review research on how humans address the simplest non-trivial perception problem, image classification, and sketch a corresponding \aip~architecture for this task. We argue that this architecture can provide a systematic way of addressing the limitations of \aip~using DNNs and an approach to assurance of human-level performance and beyond. We conclude by discussing what components of the architecture may already be addressed by existing work and what remains future work.   
\end{abstract}

\section{Introduction}
Artificial intelligence-based perception (\aip) using deep neural networks (DNN) has achieved remarkable performance. Yet as news reports can attest, \aip~can fail in surprising and catastrophic ways. This highlights the fact that, currently, the level of safety assurance possible for \aip~is insufficient to support the high levels of autonomy required for fully automated driving systems (ADS). In contrast, although human perception is imperfect, a status quo assumption by society is that the perception performance of a mature, unimpaired human is sufficient for the safe operation of a vehicle. Thus, achieving and assuring \aip~performance against a human baseline would be necessary for a societally acceptable ADS and therefore a worthy goal. 

In this paper, we take the position that this goal can be approached by studying how humans do perception and using this to construct a corresponding human-inspired \aip~architecture.  The idea of using humans as inspiration for \aip~is not new. Many of the techniques of AI are based on human psychology or neurophysiology and this trend has accelerated in recent times (e.g., ~\cite{suchan2021commonsense,malowany2020biologically,  yildirim2020efficient}). Instead, our focus is specifically on how to use the connection to humans to support the goal of \emph{safety and its assurance}. 

To investigate this concretely, we consider the basic perception task of \emph{object image classification}: does an image X depict a member of a given class C? We show that a human-inspired \aip~architecture for this task can assurably address key limitations of current DNN-based \aip~approaches while still leveraging the strengths of DNNs. 
% We also identify some essential properties of the components. We also show that the architecture has some additional desirable properties such as support for self-supervised learning.      

The remainder of the paper is structured according the following contributions: 1) we review research from the cognitive sciences on human object image classification; 2) we present a safe \aip~architecture aligned with this work; and 3) we provide justification for the architecture from various perspectives including feasibility and assurability. Finally, we give conclusions.

\section{How Humans (Probably) Do Classification}
In this section, we review research from the cognitive sciences relevant to how humans do object image classification. 
\subsection{Dual Process Models}\label{sec:dual}
In the cognitive sciences, a \emph{dual process} model of cognition is the dominant view~\cite{epstein1994integration,kahneman2011thinking,evans2013dual}. Type 1 thinking is fast, non-conscious
holistic, intuitive, and the same across different individuals. Type 2 thinking is slow, conscious, sequential conceptual reasoning that varies across individuals and is correlated with intelligence measures.
%and similar even across different species.
% Kahneman wrote a popular book describing this~\cite{kahneman2011thinking}. 

% Type 1: 
% System/Type 2: Slow, conscious Sequential, reasoning Varies across individuals, correlated with intelligence measures

The dominant view on how the two types interact is \emph{default-interventionism}~\cite{kahneman2011thinking,evans2013dual}: the Type 1 process always produces some default response quickly, and the Type 2 process intervenes to produce a potentially different response only if ``difficulty, novelty, and motivation combine to command the resources of working memory''~\cite{evans2013dual}. 
% These are conditions for triggering Type 2 process. [what about time? what about risk? - should be connected to motivation] 
The Type 1 default response may be wrong---humans often act as ``cognitive misers'' by substituting a less accurate easy-to-evaluate characteristic for a harder one, leading to biases (e.g., stereotyping). 
An important metacognitive factor is the level of ``confidence'' in the default response. When people are confident, they are less likely to invoke the Type 2 process~\cite{thompson2011intuition}. Thus, low confidence is a key triggering factor for Type 2 intervention. 

Time and risk play important roles. For fast (Type 1) binary perceptual decisions (less than 1,500\,ms), research supports the idea that evidence accumulates over time until a threshold is reached and a decision is made. In addition, there is a speed/accuracy tradeoff. If speed is a priority then accuracy may be lower, while a focus on high accuracy slows the decision (e.g.,~\cite{ratcliff2008diffusion}). 

A natural criterion for choosing the priority is the perceived risk associated with the decision. Safety-critical decisions that must be made quickly, e.g., an object appears suddenly in front of the vehicle, prioritize speed. In this case, accuracy may suffer, and Type 2 intervention is not an option, because it is slow. This suggests that even inaccurate Type 1 decisions should be appropriately conservative to manage risk. For example, if there is not enough time to determine whether the object that suddenly appeared is a pedestrian or a cyclist, a safe response may be to assume that it is a pedestrian, since this suggests a more conservative behaviour. 

Although this risk managing approach to Type 1 classification seems intuitive and prudent, it is difficult to support from research. The research on time pressure and human decision risk is focused on gambling contexts where it has been observed that when time pressure forces Type 1 decisions, these may be riskier decisions than if more time was available (e.g., ~\cite{madan2015rapid}). Since the risk in gambling contexts is measured in monetary terms, it is not clear how well these results transfer to fast safety-critical decision making.

A related well-researched area is the ``choking-under-pressure'' phenomenon exhibited by humans in high-stakes situations such as sporting events~\cite{yu2015choking}. One explanation proposed for this is that pressure induces people to consciously monitor their behaviour causing a switch from automatic and efficient (Type 1) behaviour to a slower controlled (Type 2) behaviour~\cite{baumeister1984choking}. 
% It is difficult to find research on the effects of time pressure for safety-critical decisions --- perhaps because it is too difficult to set up experimental contexts for this.  

% [can I get support for this! When there is little perception time there is greater uncertainty. There is evidence that in ultra fast categorization, superordinate categorization occurs first and this is because there is limited perceptual information. Basic level requires more time. A superordinate category is under-classification. Is there evidence that when there is uncertainty, a more conservative decision is made? ] There is evidence from studies involving fast financial decisions that faster decisions are also riskier (as measured in terms of potential financial loss).  Future high risk (or lack of progress, discomfort, etc.) motivates need for Type 2 intervention to potentially improve on Type 1 default. Uses reasoning to improve classification accuracy as more time invested~\cite{??}. 

\subsection{Object Image Classification}
Specific to the object image classification task, two prominent lines of research from different perspectives are \emph{object recognition}, studied in the neuropsychology of vision~\cite{dicarlo2012does}, and \emph{object categorization}, studied predominately in cognitive psychology and cognitive linguistics~\cite{goldstone2018categorization}.  
%(See~\cite{palmeri2004visual} for a comprehensive review of the connections between these research directions). 
Object recognition concerns the ability to assign labels to particular objects sensed by the retina, including precise identifying labels and coarser category labels. Object categorization is the more general cognitive process of grouping objects based on similar or shared features~\cite{goldstone2018categorization}. Note that the term ``categorization'' used in the cognitive sciences is synonymous with ``classification'' as used in AI contexts. 

Vision processing in the brain has two major streams: the ventral stream is responsible for object recognition, whereas the dorsal stream is responsible for visually guided action.  Recent research provides strong evidence that some Type 1 representation of a category is already in the ventral stream, expressed in terms of visual features, even though it is ultimately coded using more abstract (i.e., conceptual) features (Type 2) in downstream parts of the brain~\cite{bracci2017partnership}. The categorization in the ventral stream is fast, with a response time as little as 250-290\,ms for some categories, confirmed by multiple studies~\cite{fabre2011characteristics}.

Humans are effective at recognizing objects under different confounding visual conditions, such as varying positions to the object, lighting, context, occlusion, etc. A key function of the ventral stream is to facilitate this ability by \emph{transforming object images into representations invariant to these conditions} before further processing to categorize the object~\cite{dicarlo2012does}. Two theories dominate regarding the form of the invariant object representation. The structural description theory~\cite{biederman1987recognition} proposes a 3D parts-based representation, while in the view-based theory~\cite{poggio1990network}, objects are represented as a combination of a small set of particular 2D views that can be transformed to represent any other view.
In this paper, we will refer to the transformation of an object image to an invariant representation as \emph{object normalization}.

%[How much of object recognition is Type 1?]

% belief bias - the tendency to support the conclusion that is more consistent beliefs rather than the one that follows from the evidence - stereotyping

 Although object categorization can be seen to be part of object recognition, the research tradition in this area is focused on theories about \emph{concepts} --- the mental representation of a category. As such it is applicable to both Type 1 and Type 2 processes. The classical \emph{rule-based} theory of concepts extending back to Greek philosophers is that they consist of the necessary and sufficient conditions for membership in the category. This view has been much critiqued. For example, Wittgenstein observed that the requirement for a set of necessary conditions often does not hold due to presence of exceptions and famously illustrated this by attempting to find the necessary conditions for the category ``game''. It is also inconsistent with empirical evidence obtained by Rosch~\cite{rosch1973natural} that categories are graded, with some members more central or  typical than others, having more of the common features. This led Rosch to propose that concepts are \emph{prototype-based} with membership determined by degree of similarity to the prototype. Another dominant proposal supported by empirical evidence is that concepts are \emph{exemplar-based}~\cite{medin1978context}, where exemplars are specifically remembered examples of the category and membership is determined by collective similarity to all exemplars. Each approach has its strengths and weakness and more recently, the accepted view is that all of these approaches may be used in some combination~\cite{murphy2016there}.
 
 Both the prototype and exemplar approaches to object categorization depend on \emph{similarity judgement} to compare the observed object image with stored representations. Research on human similarity judgement is extensive (See~\cite{goldstone2012similarity} for a review). Four basic approaches have been proposed: \emph{geometric} using a distance measure in a continuous space, \emph{feature-based} aggregating the number of shared discrete features, \emph{alignment-based} extending the feature-based approach to include relations (e.g., part-of) between features, and \emph{transformation-based} based on the effort needed to transform one image into another.  

% [connection between type 1 and 2 for these?]
% [Rosch superordinate,basic, subordinate levels. Basic level is the highest level at which a single prototype is possible. It is the level at which people respond the fastest. So I think these should correspond to the easiest classification decisions for humans. However Thorpe shows that the fastest is at the superordinate level for certain categories such as animate/inanimate. I think this seems to be because of specially evolved processing for these in the ventral stream (Type 1)]

% [confidence = metacognition. What does the research on metacognition say about how it affects perception?]

% \begin{itemize}
%     \item Dual process view
%     \item Default interventionism
%     \item Object recognition
%     \item Object categorization
%     \item Risk assessment and risk management
% \end{itemize}

\begin{figure*}
    \centering
    \includegraphics[width=0.9\textwidth]{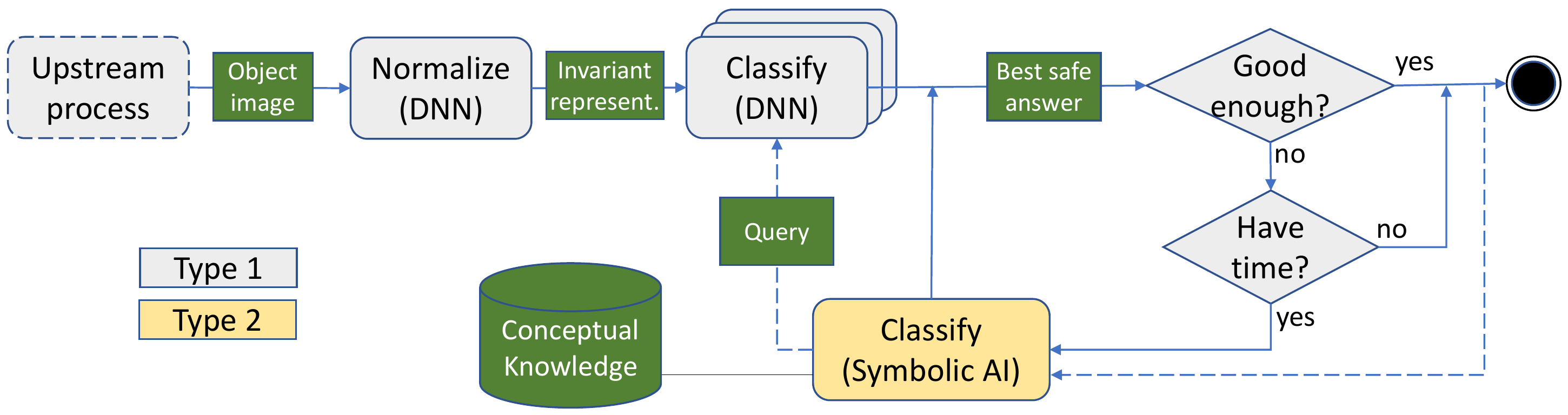}
    \caption{Human inspired classification activity diagram.}
    \label{fig:hidual}
\end{figure*}

\section{An Assurable Human-Inspired Classification Architecture}
Inspired by the research on object classification by humans presented above, in this section, we propose the high-level dual process architecture for classification shown in Fig.~\ref{fig:hidual}. Here, we assume that the fast Type 1 processes are carried out by DNNs, while slower Type 2 processes use reasoning with symbolic AI. The input to the system is an object image from an upstream process (e.g., the first stage of an object detector). In alignment with processing in the ventral stream, the first step is object normalization to eliminate the confounding effects of visual conditions. Then, a DNN-based classifier of the invariant object representation generates a classification based on visual features. We assume that these classifiers use prototype/exemplar methods to align with the human representations of concepts. Furthermore, we assume they measure confidence in their classification decision.

If the result produced by fast classifiers is inadequate (e.g., too low confidence) and if there is available time, then a reasoning process can intervene to attempt to improve the result by exploiting conceptual knowledge about object classes. The reasoning process  considers alternative classification hypotheses, then generates perceptual queries of the invariant object representation that could provide evidence to affirm or refute a hypothesis. We assume that these queries have yes/no answers and can all be viewed as classification problems\footnote{We limit the queries for simplicity, but more general Type 1 queries are be possible as well.}; thus, we ground the Type 2 reasoning process in the Type 1 perceptual process~\cite{harnad1990symbol}. Note that the a query is a recursive invocation (as indicated by the dashed arrows), since if the Type 1 process does not adequately answer a query, a Type 2 reasoning process can be invoked to intervene on it, and so on. Overall, the more time spent reasoning, the more this process can improve the quality of the classification by generating more potential hypotheses and by obtaining more evidence for hypotheses. 

\subsection{The Necessity of Dual Processes}
We may reasonably ask whether the additional complexity of a dual process approach to classification is really necessary. After all, a DNN is a universal approximator and with sufficient training examples, it should get arbitrarily accurate. However, we argue that a pure DNN approach is intrinsically limited. 

The classes of objects, such as, pedestrian, cyclist and car, that are relevant for perception by an ADS have 
% distinctive characteristics that are difficult to address using \aip~trained solely a dataset of object images. First, these classes correspond to human concepts [i.e., what must be learned by the \aip~is not an intensional definition of a set of objects but rather a representation that is behaviourally similar to the way human brains represent concepts]  
% Thus, the \aip~must learn a human concept.  the classes correspond to human concepts. 
the crucial characteristic that they are not primarily determined by visual features but rather by conceptual features. For example, something is a cyclist not because of how it looks (visual features), but because it exhibits conceptual features such as having one or more wheels, carrying human rider(s), being propelled by rider effort, etc. Assessing the presence of these features definitively may require arbitrary amounts of reasoning. This suggests that visual features are insufficient to correctly characterize these classes, and thus, a DNN trained on object images alone \emph{cannot ever achieve perfect accuracy, regardless of how many training examples are provided}.
%\footnote{Of course, memorizing all class instances is possible but this doesn't count.}. 

\begin{figure*}
    \centering
    \includegraphics[width=0.6\textwidth]{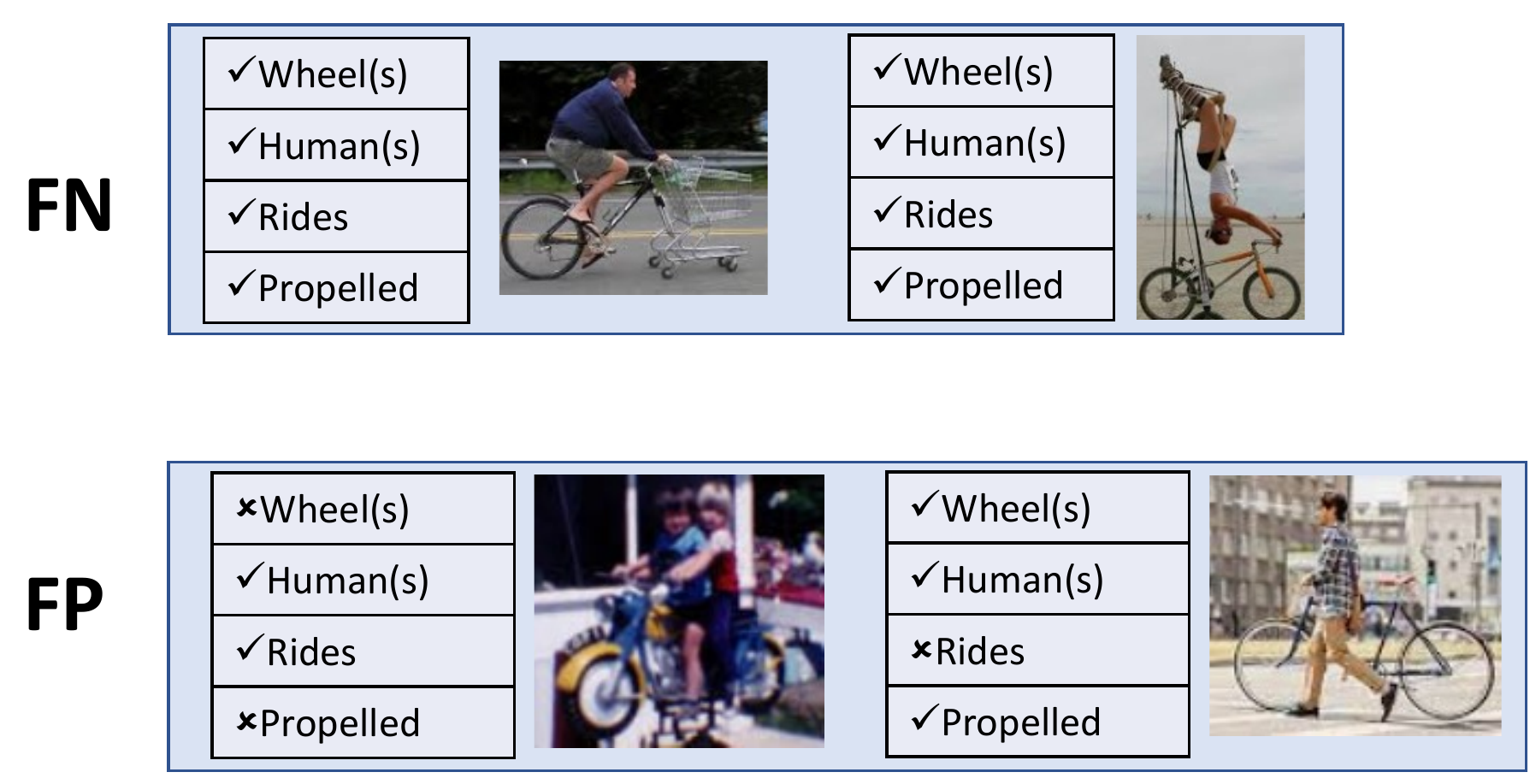}
    \caption{Visual feature based classifiers are intrinsically limited.}
    \label{fig:FNFP}
\end{figure*}

However, having a similar visual appearance for certain subsets of class instances is a common occurrence. This could be due to genetics (for ``natural kinds''), design or fashion. For example, cyclists on bicycles have visual similarity but look different from cyclists on recumbent cycles who are visually similar to each other.  When such clustering according to visual similarity is available, visual feature-based classifiers are useful approximators for these subclasses of instances. But even here, their performance is intrinsically limited as illustrated in Fig.~\ref{fig:FNFP}. It is always possible to find false negatives (FN)---unusual cyclists that fit the conceptual description but not the visual. On the other hand, we can also always find images that look like cyclists, but on careful inspection, do not satisfy the conceptual description, yielding false positives (FP). 

Despite the inaccuracies of visual-feature based classifiers, the benefit is that they may be fast in comparison to a classifier based on reasoning about conceptual features. Thus, when a safety critical decision must be made quickly, a visual-feature based classifier is preferable. This suggests that an optimal classifier strategy should follow a dual approach, leveraging visual features for speed and conceptual features for accuracy when the time is available. 
% As we see below, such a dual process strategy is consistent with research on how humans perform such object classification tasks.  
%Furthermore, in other, more intrinsically visually diverse classes such as ``chairs'', visual-feature based classifiers may be infeasible. 

To further refine this conclusion, we must address an apparent paradox. The architecture in Fig.~\ref{fig:hidual} shows that  conceptual reasoning must ultimately be grounded in visual features (or, more generally, in features of available sense modalities). This is because evidence to support conceptual hypotheses about objects in the world can only be obtained through visual means---there is no way to directly access knowledge about these objects.  Thus, all reasoning about conceptual features must be reducible to reasoning about visual features. However, if this is the case, then it would seem that \emph{visual features alone must be enough} to characterize these classes, even if they are internally encoded in terms of conceptual features.  

The way out of this apparent paradox is to acknowledge that,
while individual queries about the object image issued by a Type 2 classifier are ultimately answered using visual features, each such query appeals to potentially different visual features and the scope of such queries is limited only by the size of the knowledge base. In contrast, the set of visual features used by a Type 1 classifier for a specific class is much smaller, focused on that class only. For example, if a bicycle is decorated with flowers attached to the frame, these may create enough of a visual distortion to cause an FP in a Type 1 cyclist classifier. However, the Type 2 conceptual reasoning process can potentially identify the presence of flowers (using a Type 1 flower classifier) and conclude that these do not affect the satisfaction of the conceptual definition of cyclist. 

In this case, it is unlikely that the Type 1 cyclist classifier could ever learn to draw this conclusion because it would need to develop sensitivity to visual features about flowers. More generally, it would need to handle the visual features for every class in the knowledge base that could ever co-occur with a cyclist, which is likely to include most of the knowledge base. The dual process approach solves this scalability problem by delegating the job of ranging over the full span of world knowledge needed in the many varied, but rarer cases, to Type 2 classification and keeping  Type 1 classification focused on typical class features.

\subsection{Addressing Safety}
We assume that the safety requirements of an object classification subsystem are refined from system level (e.g., ADS) safety requirements (see~\cite{salay2021missing} for a schema of such a refinement). This refinement identifies specific performance requirements of the subsystem needed to address different potential hazard scenarios. Since these requirements are system specific, for our proposed high-level architecture we instead consider the general implications of the high-level requirement that the subsystem provides performance at least as good as humans. In particular, the following three requirements are relevant and follow from the review of human classification.

\BReq\label{req:typical}
The classification subsystem shall support accurate classification for both typical and atypical inputs.  
\EReq
Humans are able to effectively address both these types of inputs, and while it is well-known that DNN-based classifiers can achieve high accuracy on typical cases, they can often fail on unusual cases. As argued in the previous section, this is because DNN classifiers use visual features and these are only sufficient for characterizing subsets of class instances that cluster on visual similarity. These clusters identify visually prototypical class instances. However these only approximate the true class described by conceptual features, leading to both FNs and FPs for atypical cases. To correct these inevitable misperceptions by Type 1 classifiers, the architecture uses a Type 2 classifier based on conceptual reasoning. 

The decision on when to invoke the Type 2 classifier is a crucial part of the architecture (i.e., the ``good enough'' decision in Fig.~\ref{fig:hidual}). One signal relevant here is a measure of the uncertainty (or conversely, confidence) in the Type 1 classifier result. Assume the Type 1 classification process produces a categorical distribution $P(c)$ across classes and that this is \emph{calibrated}--- i.e., the value $P(c)$ for an input image accurately reflects the actual probability that $c$ is the correct class of the image. 

A true positive (TP) classification corresponds to sharp distribution with one class having high probability and the others low. A distribution close to uniform probability indicates high uncertainty and a potential FN representing a visually atypical instance (Fig.~\ref{fig:FNFP}, top). A distribution in which a few classes dominate also represents higher uncertainty indicating atypical visual ambiguity and could signal a potential FP. For example, the bottom right example in Fig.~\ref{fig:FNFP} could have highest probability for Cyclist (causing an FP) but with the probability of Pedestrian a close second. A limitation of this approach for detecting FPs is that it may require a large number of classes. For example, the bottom left example in Fig.~\ref{fig:FNFP} would only be caught if there was a class BicycleRide.

\BReq\label{req:speed}
The classification subsystem shall support classification for both fast and slow safety-critical decisions.
\EReq
This requirement acknowledges that safety-critical decisions may occur over different time-frames. For example, an object appearing suddenly ahead of the ADS requires a fast response, whereas an object causing a traffic slowdown ahead allows for a slower response. When fast classification is required, the architecture assumes that this is provided by the Type 1 process alone, since Type 2 processes are too slow. For typical cases, this can provide assurably high levels of accuracy. A limitation of the architecture is that atypical cases may be misclassified by the Type 1 classifier and this can be a safety hazard in some situations. Uncertainty measurement of the Type 1 result, as discussed above, may play a mitigating role here by signaling to the driving policy when the classification may be incorrect and a conservative action should be taken to minimize risk. 

In~\cite{salay2020purss}, a systematic way to approach this with a quantifiable safety guarantee is proposed.   
A \emph{credible set} of $P(c)$ with confidence $\alpha$, is a smallest subset of classes such that their cumulative probability is not less than $\alpha$. Because the classifier is calibrated, the true class is in the credible set with probability at least $\alpha$. Thus, if the Type 1 classifier sends the credible set as its result to the driving policy, any action it produces that is safe for all the classes in the set will be safe at least $\alpha\times 100$\% of the time. 

For example, consider the bottom right image of a person walking their bicycle in Fig.~\ref{fig:FNFP}. Assume the Type 1 classifier returns the categorical distribution $\{$Cyclist:$0.55$, Pedestrian:$0.40$, Car:$0.05\}$. Simply selecting the class with the maximum probability would make the result Cyclist, but this is an FP---the true class is Pedestrian. However, we can be 95\% sure that the true class is one of $\{$Cyclist, Pedestrian$\}$, which is the credible set for $\alpha=0.95$. If the driving policy chooses an action that is safe for both classes in this set, it will be safe at least 95\% of the time. The cost is a potentially overly conservative action. A limitation of this approach is that it requires there to exist an action that is safe for every class in the credible set, which may not always be the case. 

\BReq\label{req:visual}
The classification subsystem shall support accurate classification in the presence of confounding visual conditions within the range tolerated by humans.  
\EReq
Humans are effective at ignoring conditions such as varying positions to the object, lighting, context, occlusion, etc. However, these kinds of variations have proven to be challenging for DNN-based classifiers and are the basis for many kinds of adversarial attacks. The architecture addresses this issue by introducing the object normalizer. The Type 1 classifiers operate on the invariant representation in which the confounding effects are mostly removed. 

However, since the impact of confounding visual conditions is to introduce aleatoric uncertainty into the image, the effectiveness of object normalization and subsequent classification is ultimately limited by the amount of aleatoric uncertainty present. For example, a certain amount of lighting variation can be removed, but as the light gets lower, information loss increases until it is too great to discern the object in the image. There is, therefore, a limited range of tolerable visual conditions for both humans and machines. We require that the object normalizer + classifier combination operate at least within the  human range. Methods for eliciting formal requirements representing such human-tolerable ranges have been recently proposed~\cite{hu2020towards, hu22}.       

\subsection{Type 1/Type 2 Consistency}
We should expect that some \emph{consistency} relation holds between the Type 1 and Type 2 classifications, but what should it be? 
As discussed above, the Type 1 classification based on visual features is inherently limited---it may achieve high accuracy for typical cases but often produces FNs and FPs for atypical cases. 
Furthermore, recall that for humans, the interaction between the Type 1 and Type 2 processes is not a decision fusion of redundant perceptual processes, but rather that the Type 2 process intervenes to improve on the Type 1 result when necessary and possible. This relationship is inherited by the proposed architecture. Thus, the Type 2 classification is considered both to be \emph{authoritative} and it must be \emph{no worse} than the Type 1 classification. The latter condition suggests that when Type 1 is TP, then so must Type 2, but when Type 1 is FN or FP, Type 2 may be the same or TP. 

Note that we do not assume the Type 2 classification is necessarily always TP even though it is considered authoritative, since its accuracy is still limited when excessive aleatoric uncertainty is present. Furthermore, the degree of improvement over the Type 1 classification is limited by the reasoning time available, richness and correctness of the conceptual knowledge base and accuracy of the Type 1 classifiers used to answer queries. 

Until now we have been discussing the kind of \emph{classification consistency} that must hold between the Type 1 and Type 2 classification. Another kind of consistency is \emph{risk consistency}---how is the safety of the classifications related? If we assume that a correct classification is always at least as safe (i.e., leads to a driving policy action that is not more hazardous) as a misclassification, then our classification consistency requirement implies that, when time is not a safety-critical factor, the Type 2 classification is always at least as safe as the Type 1 classification. 

However, not all misclassifications are unsafe. For example, misclassifying a pedestrian as a cyclist, when it is still far ahead, may not lead to different behaviour by an ADS. Thus, the hazardousness of a given misclassification is situation-dependent. Can this fact be exploited to produce a stronger risk consistency requirement? In an assurance case, a fine-grained analysis of hazardous patterns of misperceptions relevant in different driving scenarios can provide a correspondingly fine-grained and risk-aware set of performance requirements for the Type 1 classifiers~\cite{salay2021missing}. Such a set of requirements identify the kinds of images that are more likely to cause hazardous actions if misclassified, thus the training of Type 1 classifiers can focus more on these. 

Another approach to stronger risk consistency is based on the credible set approach to representing uncertainty discussed above. If a Type 1 classifier produces the credible set for a required level of confidence $\alpha$ as output, then even though uncertainty is present, a driving policy can still perform a safe action, if one exists. In limited operational design domains, it may be possible to show that a safe action exists in every situation for any subset of classes. Thus, in such a restricted context we can satisfy the following additional risk consistency condition: the Type 1 classification will always be as safe as the Type 2 classification $\alpha\times 100\%$ of the time. Note that, even here, a Type 2 classification is still preferable when time permits because the action based on an uncertain Type 1 classification is more conservative than necessary and may hamper other ADS objectives such as progress or comfort.

\section{Validation}
Although the proposed architecture is human-inspired, this alone is not sufficient to justify it. In this section, we validate the architecture by analyzing the feasibility and assurability of the components.

\subsection{Feasibility of Architecture Components}
We briefly review existing work that could address the requirements of architecture components.
\subsubsection{Object Normalization}
The field of computer graphics studies how to render object and image-taking specifications (e.g., 3D mesh, light sources, textures, camera position, etc.) into an object image. The problem of \emph{inverse graphics} is how to produce such a specification from an object image; thus, it performs the task of object normalization. Solving the inverse graphics problem is active research and various recent approaches using neural networks have been proposed (e.g.,\cite{yao20183d,deng2019cerberus,yildirim2020efficient}). The idea of \emph{capsule networks} is a prominent approach~\cite{HintonSF18} where the network learns an object class by decomposing into object parts and their structural relationships.

Another field that is relevant here is \emph{embodied AI}~\cite{chrisley2003embodied}. Humans learn about objects by engaging with them directly in the world. In this way, they automatically learn what aspects of their experience are irrelevant to tasks such as classification (e.g., their position relative to the object or the orientation of the object).  Artificial agents may be able to obtain these same benefits if they learn in a similar way rather than being trained from a predefined dataset of static images~\cite{smith2005development}. To facilitate this, various simulation tools have been developed that allow artificial agents to roam and interact in a simulated world to learn about it directly~\cite{duan2021survey}. An example of applying this to object normalization is to learn spatial invariants of object classification by approaching objects in different ways in the simulated world~\cite{caudell2011retrospective}. 

\subsubsection{Type 1 Classification} An emerging trend for DNNs is \emph{dynamic} inference where the DNN can exit early if needed~\cite{teerapittayanon2016branchynet}. This can implement the speed/accuracy tradeoff observed in the ventral stream. 

Classifiers that use DNNs are typically structured as a series of convolutional layers followed by fully connected layers. The lack of interpretability of these approaches limits their applicability as a Type 1 classifier when safety assurance is required. Alternative and interpretable DNN architectures based on prototype or exemplar approaches have recently been investigated and have shown positive results (e.g., ~\cite{li2018deep,hase2019interpretable, papernot2018deep}). The paper ``This looks like that: deep learning for interpretable image recognition''~\cite{chen2018looks} is good example of such architectures. Here, a classifier for different bird species is developed by learning for each class a set of prototypical image fragments taken from training images. Inference is then done by judging similarity of the learned prototypes to an input image and assigning the image to the class with the best fit.

%[the issue with similarity]

\subsubsection{Type 2 Classification} In the absence of general AI, assurance can benefit from a wide range of classical AI approaches, which need to focus on \emph{explaining} what the object image is. Thus, approaches to abductive reasoning are applicable. As discussed above, the classes used by ADSs often do not possess a common set of necessary conditions; thus, traditional monotonic logics may be inappropriate. Non-monotonic logics (e.g., default logic) have been developed to express class membership rules which allow exceptions. Case-based reasoning aligns well with exemplar-based categorization. Description logics have concepts as first class entities and have been extended to support prototype-based reasoning (e.g.,~\cite{baader2016reasoning}). Reasoning using formalized ``commonsense'' theories provides a way to utilize human conceptual knowledge about various domains (e.g., physics of objects)~\cite{davis2017logical,suchan2021commonsense}.
Another line of research relevant here concerns formal executable models of conceptual categorization such as Dual PECCS~\cite{lieto2017dual} that incorporates both prototype and exemplar based reasoning.   
%[Say something about commonsense reasoning?]
Finally, integration with Type 1 classification DNNs can use neuro-symbolic approaches~\cite{neurosym21}, information fusion frameworks~\cite{suchan2021commonsense}, and imposing causal models on neural networks~\cite{Geiger:Wu:Lu-etal:2021}.

\subsection{Safety Assurance}
% In this section, we address the central theme of this paper - that an architecture presented in Fig.~\ref{fig:hidual} aligned with how humans do classification, can provide a greater opportunity for safety assurance of human performance baseline. We assess the assurability of each component in the architecture but we begin with some general observations about performance comparisons.
% While the architecture presented in Fig.~\ref{fig:hidual} is high-level, to do this, we briefly consider existing work that could be applicable to implement that various components (it is beyond the scope of this paper to provide component details).

\subsubsection{Performance Comparison}
An assurance argument regarding a human baseline must rely on some performance metrics for comparing component performance to the baseline. A naive way to proceed is to use one of the many performance metrics that have been proposed to compare the performance of different classifiers (e.g., accuracy, precision, F1-score). Such ``generic'' metrics are problematic for several reasons. First, such comparisons should be ``species-fair'' and not be biased by operational differences ~\cite{firestone2020performance}. For example, the retina is high resolution in the fovea but loses resolution and is color blind at the periphery. Thus, it sees an image differently than a DNN that gets an image as a uniform pixel grid. This difference can result in different classification accuracy of an image even if this has nothing to do with classification knowledge.

Second, comparisons should be \emph{risk-aware}---performance differences in a context that is not safety relevant are not important. One way to achieve this is to define specialized perception performance metrics for different hazardous driving scenarios~\cite{salay2021missing}. Finally, 
because generic metrics average performance over many trials, an \aip~may obtain the same value as a human on the metric but still make, what to humans seem like unjustifiable errors (e.g., adversarial examples), undermining the assurance argument. To address this, performance measurements should be made for different difficulty categories for humans. 
In particular, cases that are easy for humans (e.g., variations due to confounding visual conditions) should also be easy for the \aip---adversarial examples violate this condition. Furthermore, the use of an \emph{error consistency} metric is needed here, which measures the degree to which the \aip~is making the same decision as a human on individual trials~\cite{geirhos2020beyond}. A high error consistency provides evidence that the \aip~is following a similar strategy as the human in its classification decision. Note however that we are only interested in preserving strategies where humans make correct decisions and do not want to replicate their weaknesses.

\subsubsection{Object Normalization}
The object normalizer identifies where the confounding effects of visual conditions are explicitly addressed in the architecture. Thus, the assurance argument regarding robustness to adversarial cases focuses here. Furthermore, since we take human performance as a baseline, the performance of the normalizer needs only to be assured up to human tolerable bounds on these conditions (e.g., maximum level of fog after which human performance is inadequate). Methods for eliciting formal requirements representing such bounds, as well as corresponding testing criteria, have been recently proposed~\cite{hu2020towards, hu22}.  

A generic DNN-based object normalizer would be reusable for different classification tasks allowing any assurance effort to be amortized over all its applications. Thus, although not-interpretable, it could be subjected to increased and extensive testing scrutiny. In addition, this testing effort would be robust because it is not subject to distributional shift or dependencies on community-specific norms since ``objecthood'' is such a basic concept. 

Techniques for formally verifying DNNs are being developed (e.g.,\cite{liu2019algorithms}). Thus, formal verification may be a possible solution for invariances that can be expressed formally as object image transformations (e.g., affine transformations or injected Gaussian noise). Formalizable aspects of object normalization may also allow non-data-driven implementation amenable to traditional assurance practices.

\subsubsection{Type 1 Classification}
A significant positive impact of object normalization is to simplify the classification problem since the classifier needs only to learn the visual features of the class instances in an idealized setting. This reduces the size and diversity needed in the dataset to assure adequate sample coverage of the input distribution. It also improves generalization by reducing the likelihood of spurious correlations with noncausal features of the input.   

Prototype/exemplar-based classifier approaches using DNNs provide interpretability by allowing human inspection of the prototypes/exemplars to determine whether they are meaningful. For example, in ``This looks like that'' discussed above, the prototype fragments of bird images can be inspected by birding experts to determine whether they are indicative of the classes they correspond to. This expert assessment provides evidence for correctness in the safety argument. Unlike the many post hoc explainability mechanisms that have been  proposed for DNNs, such as saliency maps, interpretability provides the faithful explanations needed for assurance~\cite{rudin2019stop}. 

Another potential benefit of prototype/exemplar-based classifier approaches is the alignment with how humans represent concepts. This could provide evidence that the classifier generalizes in the same way as humans---i.e., by judging similarity to prototypes (and/or exemplars) that have been validated as conforming to community or expert opinion. However, the validity of this ``evidence from alignment'' argument depends also on the alignment of the similarity metric used with how humans judge similarity. If generic object similarity judgement can be learned by a DNN, then, like an object normalizer, this is could be a reusable component that can be given a higher degree of testing scrutiny. However, there is mixed evidence about whether this is possible. Earlier studies show comparable performance for DNN-based similarity judgment relative to humans, but a more recent study found that DNNs cannot outperform humans when more complex categorical knowledge is needed to judge similarity~\cite{jozwik2017deep}. This suggests that similarity judgement may itself be a perception task that requires a dual process treatment to achieve human-level performance.

%[Bounding the occurrence of hard subtle cases] 

\subsubsection{Type 2 Classification}
The knowledge base used by reasoning here is expressed in terms of human understandable concepts; therefore, it is interpretable and inspectable. This allows verification of alignment with community-specific consensus knowledge about object classes. 
Additionally, since reasoning is formal and based on a logic,
% logic (although these could be non-traditional logics including non-monotontic, probabilistic, fuzzy, etc.)
evidence of internal consistency (i.e., soundness) and areas of (in)completeness of the knowledge base can be facilitated using formal methods.

The requirement of classification consistency imposes an important constraint between the knowledge at the Type 1 and Type 2 levels that must be verified as part of an assurance argument. Automatic cross-validation methods between the levels could facilitate this. For example, Type 2 reasoning could be used to label images with semantic information that is then used to train or test the Type 1 classifiers. 
%This may reveal new relevant features and feature dependencies that should inform how to . 
Reasoning about the scope of conceptual knowledge used by Type 2 could form the basis for completeness claims about the Type 1 classifiers and the datasets used to train and test them.

% \begin{itemize}
%     \item Go through the assurance slides
%     \item performance comparisons of \aip~to humans should be measured according to human-based ``hardness'' 

% \end{itemize}

% \section{Related Work}
% \begin{itemize}
%     \item Other human-inspired dual process approaches
%     \begin{itemize}
%         \item  Goyal/Bengio
%         \item Suchan
%         \item \cite{malowany2020biologically}
%     \end{itemize}
%     \item Other human baseline work
%     \begin{itemize}
%         \item ADS baselines about accidents?
%         \item 
%     \end{itemize}
%     \begin{itemize}
%         \item CV-HAZOP
%     \end{itemize}
% \end{itemize}

\section{Related Work}

Dual-processing architectures have been proposed for robotics (e.g., \cite{dual09}) and AI (e.g., \cite{fastSlowAI2021}) but rarely discussed from the assurance perspective. The closest work is by Jha et al. ~\cite{model-centered21}, who also advocate for such architectures to support assurance. They propose that autonomous systems use (i) Type 1 processing to construct their world models and predict future observations and (ii) Type 2 processing to refine or revise the models when the prediction errors (aka surprise) become high. They target fusing observations from multiple sensors over time (sequential Bayesian filtering) and propose that an assurance argument focuses on (i) guarding against systematic misperceptions by Type 1 processing that fail to generate surprise and (ii) the safe handling of surprise by Type 2 processing. Our contribution is complementary: we explore image classification in depth, emphasizing the distinction between visual and conceptual features, the importance of normalization, and the role of time available to make a decision.
Whereas surprise is an important trigger for Type 2 processing, so are other types of uncertainty, including ambiguity and novelty~\cite{novelty-surprise13}.

\section{Conclusion}
Although imperfect, human perception performance is often assumed to serve as a minimum baseline for safety that a societally acceptable \aip~must meet. However, it is widely known that while current state-of-the-art \aip~has achieved high levels of performance using DNNs, they still fall short of this baseline.  In this paper, we review research on how humans do the basic perception task of object classification. Then we propose a dual process architecture for a safety assurable object classification \aip~aligned with the findings of this research. We discuss how such an architecture is both potentially feasible and assurable. 

We plan on investigating several issues as part of future work. First, while this paper explores a dual processing architecture for classification, the ideas must be further developed for more general perception and decision making, potentially in a unified way. This should also go beyond a single modality like vision. When a fast and critical decision needs to be made, one may need to introduce additional sensing modalities. For example, tailpipe fumes on a cold day may appear in LiDAR like a potentially solid object, but a camera image can easily remove this ambiguity. Second, an interesting next step would be to develop a safety argument template that could be evolved and drive the development of concrete \aip~architectures in a safety-first manner. Finally, a key limitation is still the challenge to be robust to and detect out-of-distribution (OOD) samples at the Type 1 level when it needs to be fast and we intend to explore this further (plus validating the hypothesis that Type 2 can refute Type 1 for OOD samples in the long run with sufficient accuracy). Perhaps neuroscience can be helpful here too by providing insights into how the brain deals with uncertainty and novelty.  Ultimately, the lessons we can learn from the human brain may be the key to achieving assurable and societally acceptable \aip.  

%[Are DNN weakenesses addressed?]

\bibliography{references}

\end{document}

%% file: macros.tex
\newcommand{\aip}{AIP}

\newtheorem{mymethod}{Method}
\newtheorem{theorem}{Theorem}
\newtheorem{mydefinition}{Definition}
\newtheorem{myobservation}{Observation}
\newtheorem{myremark}{Remark}
\newtheorem{myproposition}{Proposition}
\newtheorem{myclaim}{Claim}
\newtheorem{mylemma}{Lemma}
\newtheorem{mycorollary}{Corollary}
\newtheorem{myexample}{Example}
\newtheorem{myexamples}{Examples}
\newtheorem{myalgorithm}{Algorithm}
\newtheorem{myconstruction}{Construction}
\newtheorem{myrule}{Rule}
\newtheorem{myreq}{Requirement}

\newcommand{\bolddot}{\hspace{-1.5mm}\textbf{.}\ \  }

\newcommand{\BT}{\begin{theorem}}
\newcommand{\ET}{\end{theorem}}
\newcommand{\BCR}{\begin{mycorollary}\bolddot}
\newcommand{\ECR}{\end{mycorollary}}
\newcommand{\BPR}{\begin{myproposition}}
\newcommand{\EPR}{\end{myproposition}}
\newcommand{\BL}{\begin{mylemma}}
\newcommand{\EL}{\end{mylemma}}
\newcommand{\BCM}{\begin{myclaim}}
\newcommand{\ECM}{\end{myclaim}}

\newcommand{\BD}{\begin{mydefinition}}
\newcommand{\ED}{\end{mydefinition}}
\newcommand{\BEX}{\begin{myexample}}
\newcommand{\EEX}{\end{myexample}}
\newcommand{\BEXS}{\begin{myexamples}}
\newcommand{\EEXS}{\end{myexamples}}
\newcommand{\BOB}{\begin{myobservation}}
\newcommand{\EOB}{\end{myobservation}}
\newcommand{\BR}{\begin{myremark}}
\newcommand{\ER}{\end{myremark}}
\newcommand{\BAL}{\begin{myalgorithm}}
\newcommand{\EAL}{\end{myalgorithm}}
\newcommand{\BAM}{\begin{mymethod}}
\newcommand{\EAM}{\end{mymethod}}

\newcommand{\BCO}{\begin{myconstruction}}
\newcommand{\ECO}{\end{myconstruction}}
\newcommand{\BRule}{\begin{myrule}}
\newcommand{\ERule}{\end{myrule}}
\newcommand{\BReq}{\begin{myreq}}
\newcommand{\EReq}{\end{myreq}}